# Compressed Sensing Parallel MRI with Adaptive Shrinkage TV Regularization

Raji Susan Mathew, and Joseph Suresh Paul

Medical Image Computing and Signal Processing Laboratory, Indian Institute of Information Technology (IIITM-K), Kerala, India.

*Abstract*— Compressed sensing (CS) methods in magnetic resonance imaging (MRI) offer rapid acquisition and improved image quality but require iterative reconstruction schemes with regularization to enforce sparsity. Regardless of the difficulty in obtaining a fast numerical solution, the total variation (TV) regularization is a preferred choice due to its edge-preserving and structure recovery capabilities. While many approaches have been proposed to overcome the non-differentiability of the TV cost term, an iterative shrinkage based formulation allows recovering an image through recursive application of linear filtering and soft thresholding. However, providing an optimal setting for the regularization parameter is critical due to its direct impact on the rate of convergence as well as steady state error. In this paper, a regularizer adaptively varying in the derivative space is proposed, that follows the generalized discrepancy principle (GDP). The implementation proceeds by adaptively reducing the discrepancy level expressed as the absolute difference between TV norms of the consistency error and the sparse approximation error. A criterion based on the absolute difference between TV norms of consistency and sparse approximation errors is used to update the threshold. Application of the adaptive shrinkage TV regularizer to CS recovery of parallel MRI (pMRI) and temporal gradient adaptation in dynamic MRI are shown to result in improved image quality with accelerated convergence. In addition, the adaptive TV-based iterative shrinkage (ATVIS) provides a significant speed advantage over the fast iterative shrinkage-thresholding algorithm (FISTA).

## I. INTRODUCTION

The duration of data acquisition in magnetic resonance (MR) imaging is constrained by the considerations of patient discomfort, motion artefacts, and patient throughput. One approach to speeding up the acquisition is to reconstruct the final image from a limited number of k-space samples. Recent advances in the application of compressed sensing (CS) theory to MR image reconstruction require the signal to have a sparse representation derived from incoherent measurements. Since randomly acquired MR data with incoherent aliasing can be conveniently represented in a sparse domain using wavelets or finite differencing, CS principles can be directly applied for efficient MR image recovery.

Acquisition of k-space samples at a rate lower than the Shannon-Nyquist sampling limit renders image recovery an ill-posed problem. Various algorithms have been proposed to tackle such ill-posed problems, including non-linear reconstruction procedures that enforce data consistency along with sparsity promoting regularization [1]. In comparison to the frequency localization transforms for sparse domain representation [2-6], TV-based sparsity-promoting regularization [7-9] appears to offer potentially superior edge preservation and structure recovery, while suppressing the noise present in the image [10]. The responsible property is the piecewise-smooth nature of the TV regularization term. However, the major difficulty hindering TV based reconstruction in finding a fast numerical solution is the non-differentiability of its cost term.

Numerous algorithms have been proposed to address the non-differentiability problem. In an approximation to the TV model proposed by Rudin, Osher and Fatemi (ROF model), an artificial time marching scheme was introduced to solve the Euler–Lagrange equation (ELE) [7]. To overcome the slow convergence, Vogel and Oman proposed [11] a fixed point iteration scheme in which the ELE equation is linearized by lagging the diffusion coefficient, and the update obtained by solving a sparse linear equation. In an attempt to improve the speed, Chan, Golub and Mulet [12] proposed the primal dual idea by treating the normalized gradient term in the ELE as an independent variable, with a pair of nonlinear equations solved using Newtons method. Later, Chambolle [13] proposed a semi-implicit gradient descent algorithm based on the Fenchel dual formulation for the TV denoising problem, and primal-dual proximal splitting algorithm for solving generalized non-smooth composite convex minimization problems [14]. An alternating direction method based on variable splitting was also proposed by Wang *et al.* [15] for efficient TV regularized image recovery. An alternative variable splitting method using Bregman iterations was proposed by Goldstein and Osher [16] to address $l_1$-regularized optimization problems. Implementation of TV based minimization using alternating direction method of multipliers (ADMM) was applied to image restoration by Ng *et al.* [17] for Gaussian noise corrupted images. In multiplicative noise models, ADMM for TV regularization were used with cost functions based on penalized likelihood by Carlavan and Blanc-Féraud [18], and Kullback-Leibler (KL) divergence by Teuber *et al.* [19]. An overview of these different approaches is provided in Burger *et al.* [20] and Glowinski *et al.* [21].

Selecting an optimal setting for the regularization parameter in TV optimization plays a central role in determining its reconstruction accuracy. To this end, several parameter selection strategies have been applied to TV optimization such as the discrepancy principle [22-25], the generalized cross-validation method [26], the unbiased predictive risk estimator (UPRE) [27], the variational Bayesian approach [28, 29], and the majorization-minimization (MM) approach [30].

In recent years, a number of schemes have been proposed to adapt the regularization parameter based on the discrepancy principle, which assumes that either the noise variance or the signal-to-noise ratio (SNR), is known a priori. The efficiency of these adaptation strategies relies completely on prior knowledge of the noise variance. To overcome this limitation, we propose an iteration-dependent parameter update scheme along the lines of the generalized discrepancy principle (GDP), which does not require precise knowledge about noise present in the MR data. The adaptation strategy is an extension of our previously established update scheme for the wavelet threshold [31] to the TV iterative shrinkage (TVIS) framework recently proposed by Michailovich [32], in which the method of iterative shrinkage is directly applicable to the TV functional rather than its smoothed versions. As the method is an adaptive version of TVIS, we refer to the proposed scheme hereinafter as adaptive TVIS (ATVIS). Accordingly, TV based image recovery is achieved through a recursive application of two simple procedures, linear filtering and soft thresholding, as in the case of wavelet based regularization.

In the proposed adaptive strategy, we draw a parallel between (1) the residual error versus the consistency error, and (2) the perturbation error versus the difference between coefficients before and after soft thresholding in the gradient domain (i.e., the sparse approximation error). The adaptation relies on the fact that the absolute difference between TV norms of the consistency and sparse approximation errors, referred to as the discrepancy level, approaches zero in the absence of noise and shows up as a small noise-dependent positive quantity at convergence. An adaptive strategy is enforced by using a discrepancy level dependent function to update the threshold. In the following, we exploit the dependence between the extent of denoising and data consistency, to adapt the threshold towards accelerated reduction of the steady state error with a reduced loss of resolution. We also show that the proposed adaptation scheme is easily applicable to multi-channel in vivo data acquired with different sampling patterns, and demonstrate significant improvements in both speed and image quality. Moreover, the adaptation scheme is shown to be suitable for accelerating dynamic MR image reconstruction, in which the temporal gradient (TG) sparsity is also used as an auxiliary constraint for improved reconstruction. This is achieved by adaptively varying the TG threshold to enhance the TG sparsity, and accelerate the convergence to lower steady state reconstruction errors.

This paper is organized as follows-: Section II provides a brief overview of the TVIS algorithm, threshold adaptation and the algorithmic procedure. Results pertaining to ATVIS in both single channel and multi-channel in vivo data are presented in Section III. Reconstructions obtained by applying the adaptation procedure to dynamic MRI are also included. A discussion relating to the implementation of ATVIS, and a concluding summary, are provided in Sections IV and V, respectively.

## II. THEORY

*A. Reconstruction Model*

The undersampled k-space representation $K_u$ of an image $U$ is given by

$$K_u = F_u U, \qquad (1)$$

where $F_u$ denotes the undersampled Fourier operator obtained by setting the unacquired frequency points to zero. Recovery of the image $U$ from (1) is ill-posed due to following reasons: i) the system of equations is underdetermined, ii) the required solution is unstable due to the numerically ill-conditioned Fourier operator, and iii) the presence of acquisition noise. With prior knowledge that the signal is sparse, recovery involves seeking the sparsest amongst all the solutions of (1). Although this gives rise to an $l_0$-problem which is NP-hard [33], practical applications rely on a convex relaxation approximation based on $l_1$-norm minimization. With $\Psi$ representing the sparsifying transform operator, the regularized reconstruction for a given consistency measure $\varepsilon$ is obtained as

$$\min_U \|\Psi U\|_1 \text{ subject to } \frac{1}{2}\|K_u - F_u U\|_2^2 \leq \varepsilon, \qquad (2)$$

where $\|\Psi U\|_1$ is the sparsity constraint and $\|K_u - F_u U\|_2^2$ is the data consistency constraint based on the $l_2$-norm. In the unconstrained form, the optimum solution is determined by minimizing

$$\min_U \frac{1}{2}\|K_u - F_u U\|_2^2 + \beta \|\Psi U\|_1, \qquad (3)$$

where $\beta$ represents the tradeoff between sparsity and data fidelity. In TV regularized reconstruction, the sparsifying transform $\Psi U$ denotes the gradient image $\boldsymbol{d}$. The total variation $\|\boldsymbol{d}\|_1$ can be defined isotropically or anisotropically as

$$\|d\|_1 = \begin{cases} \sum_{ij} \left((\nabla_x U_{ij})^2 + (\nabla_y U_{ij})^2\right)^{\frac{1}{2}}, & (\text{Isotropic}) \\ \sum_{ij} |\nabla_x U_{ij}| + |\nabla_y U_{ij}|, & (\text{Anisotropic}) \end{cases}, \quad (4)$$

where $\nabla_x$ and $\nabla_y$ represent the forward finite difference operators on the horizontal and vertical coordinates. The definition of gradient $\nabla$ depends on the boundary conditions. With symmetric boundary conditions (SBC),

$$(\nabla U)_{n,m} = \begin{cases} (U_x)_{n,m} = U_{n,m} - U_{n-1,m} \text{ with } U_{-1,m} = U_{0,m} \\ (U_y)_{n,m} = U_{n,m} - U_{n,m-1} \text{ with } U_{n,-1} = U_{n,0}, \end{cases} \quad (5)$$

where $n = 0, 1, \ldots, N-1$ and $m = 0, 1, \ldots, M-1$. In the case of periodic boundary conditions (PBC), the gradient is defined as

$$(\nabla U)_{n,m} = \begin{cases} (U_x)_{n,m} = U_{n,m} - U_{n-1,m} \text{ with } U_{-1,m} = U_{N-1,m} \\ (U_y)_{n,m} = U_{n,m} - U_{n,m-1} \text{ with } U_{n,-1} = U_{n,M-1}. \end{cases} \quad (6)$$

A solution to the optimization problem (3) is provided in the following section.

*B. Derivative Shrinkage*

An iterative shrinkage approach to TV is based on the assumption that both the original and observed images belong to a bounded norm vector space $\mathbb{U}$. The bounded norm vector space require all the vectors in $\mathbb{U}$ to be bounded and, hence, possess a finite $l_2$-norm. Furthermore, elements of $\mathbb{U}$ are also constrained to have zero mean value, which leads to a formal definition of $\mathbb{U}$ as

$$\mathbb{U} = \{U \in \mathbb{R}^{N \times M} | \langle \mathbf{1}, U \rangle = 0, \|U\|_2 < \infty\}, \quad (7)$$

where $\mathbf{1}$ denotes an $N \times M$ matrix of ones. Image recovery in TV minimization is achieved using a left inverse operator $\mathcal{X}(d)$ given by

$$\mathcal{X}(d) = \mathcal{F}^{-1}\left(\mathcal{F}(div(d)) \odot W_i\right), \quad (8)$$

where $\odot$ denotes the Hadamard product, $div(\cdot)$ the discrete divergence operator, $\mathcal{F}$ represents the discrete cosine transform operator in the case of SBC and the discrete Fourier transform operator, otherwise. Also, $W_i$ denotes the integration filter coefficients defined as $(W_i)_{k,l} = (2\cos(a\pi k/M) +$ $2\cos(a\pi l/N) - 4)^{-1}$, with $a = 1$ for SBC and 2 for PBC, respectively [32]. Thus the optimization problem in (3) can now be represented in the derivative space as

$$\min_{\boldsymbol{d}} \frac{1}{2} \|K_u - F_u \mathcal{X}(\boldsymbol{d})\|_2^2 + \beta \|\boldsymbol{d}\|_1, \quad (9)$$

where the optimization problem in the image space $U \in \mathbb{U}$ is now replaced by the minimization of its corresponding gradient image $\boldsymbol{d} \in \mathbb{D}$. This is similar to the $l_1$-norm constrained reconstruction problem in [34], enabling the reconstruction using iterative soft thresholding. Therefore, the solution to the TV optimization problem can be implemented using the following steps incorporating iterative soft thresholding. In the first step, the Landweber update for the gradient image is obtained as

$$\widehat{\boldsymbol{d}}^{(k+1)} = \boldsymbol{d}^{(k)} + \Psi\left(F_u'\left(K_u - F_u \mathcal{X}(\boldsymbol{d}^{(k)})\right)\right), \quad (10)$$

where $\widehat{\boldsymbol{d}}$ and $\boldsymbol{d}$ denote the gradient images before and after thresholding. As the second term in the RHS of (10) represents the gradient image $\boldsymbol{\epsilon}_{res}$ of the consistency error, the Landweber update now becomes

$$\widehat{\boldsymbol{d}}^{(k+1)} = \boldsymbol{d}^{(k)} + \boldsymbol{\epsilon}_{res}^{(k+1)}. \quad (11)$$

Application of the soft thresholding function to $\widehat{\boldsymbol{d}}^{(k+1)}$ gives

$$\boldsymbol{d}^{(k+1)} = \mathcal{T}_\beta\left(\widehat{\boldsymbol{d}}^{(k+1)}\right), \quad (12)$$

where $\mathcal{T}_\beta$ denotes a component-wise application of the soft thresholding function applied to complex data, defined by

$$\left(\boldsymbol{d}^{(k+1)}\right)_i = \begin{cases} \left(\widehat{\boldsymbol{d}}^{(k+1)}\right)_i - \beta \frac{\left(\widehat{\boldsymbol{d}}^{(k+1)}\right)_i}{\left|\left(\widehat{\boldsymbol{d}}^{(k+1)}\right)_i\right|}, & if \left|\left(\widehat{\boldsymbol{d}}^{(k+1)}\right)_i\right| > \beta \\ 0, & if \left|\left(\widehat{\boldsymbol{d}}^{(k+1)}\right)_i\right| \leq \beta. \end{cases} \quad (13)$$

*C. Selection of the Initial Threshold*

Since the proposed threshold adaptation scheme requires a monotonically decreasing pattern, a higher initial threshold is required to achieve faster reduction to the smallest possible relative $l_2$-norm error (RLNE) [35-37]. Therefore, the threshold selection procedure provided by Michailovich [32] cannot yield best results when used with our adaptation scheme. Alternatively, the universal threshold proposed by Sardy and Monajemi [38] using finite differences is found to be more suitable in this case. The universal threshold is obtained as

$$\beta^{(0)} = \hat{\sigma} \mathcal{G}_{\mu,\gamma}^{-1}\left(1 - 2/\sqrt{\log P_M}\right), \tag{14}$$

where $P_M = 2N(N-1)$, $\mathcal{G}_{\mu,\gamma}^{-1}(p) = \mu - \gamma \log(-\log p)$, $\mu = \exp(\hat{a}^{\mu_0} + \hat{b}^{\mu_0} \log \log N)$, $\gamma = \exp(\hat{a}^{\gamma_0} + \hat{b}^{\gamma_0} \log \log N)$, with the values of the coefficients as $\hat{a}^{\mu_0} = -0.395$, $\hat{b}^{\mu_0} = 0.552$, $\hat{a}^{\gamma_0} = -1.512$ and $\hat{b}^{\gamma_0} = -0.247$ [38]. The standard deviation of the noise, used in threshold determination, is a scaled median absolute deviation of the finite differences given by [39]

$$\hat{\sigma} = \left(1.4826/\sqrt{2}\right) \cdot median(|(\hat{d}^{(1)}) - median(\hat{d}^{(1)})|). \tag{15}$$

*D. Adaptive Derivative Shrinkage*

Image recovery using iterative shrinkage approach requires knowledge of an optimal regularization parameter that depends on the presence of noise, image size, image type, etc. Although the reconstructed image quality is well-preserved near the optimal parameter value, it deteriorates rapidly away from this value. Furthermore, the procedures for finding the optimal parameter value in the absence of information on noise level and its statistics, can pose difficulties in optimal reconstruction. As a consequence, inferences obtained using different selection strategies can lead to suboptimal reconstructions. These difficulties can be overcome with the usage of an adaptive parameter.

An adaptive threshold can be realized by reducing the difference between the TV norms of the gradient domain consistency error and the sparse approximation error. The sparse approximation error is defined in the derivative domain and accounts for the perturbations resulting from noise as well as the deviation of the adopted basis from the true basis. It is given by

$$\epsilon_n^{(k)} = \hat{d}^{(k)} - d^{(k)}. \tag{16}$$

Analogous to the formulation of Mathew and Paul [31], the discrepancy level, defined as the difference between the TV norms of the consistency and sparse approximation errors, can be represented as

$$\mathfrak{d}_1^{(k)} \triangleq \left|\left\|\epsilon_{res}^{(k+1)}\right\|_1 - \left\|\epsilon_n^{(k)}\right\|_1\right|. \tag{17}$$

Following steps similar to (17) to (22) in [31], it can be easily deduced that

$$\left\|\boldsymbol{\epsilon}_{\text{res}}^{(k+1)}\right\|_1 \geq \left\|\boldsymbol{\epsilon}_n^{(k)}\right\|_1. \tag{18}$$

The above representation bears a close resemblance to the GDP, except that both the errors are expressed in terms of the TV norm. Since the TV norm is the sum of absolute gradient values, (18) can be represented as

$$\sum_i \left|\epsilon_{\text{res}}^{(k+1)}(i)\right| > \sum_i \left|\epsilon_n^{(k)}(i)\right|. \tag{19}$$

Using the definition of soft thresholding, the consistency and sparse approximation errors can be expressed as

$$\left(\epsilon_{\text{res}}^{(k+1)}\right)_i = \begin{cases} (\widehat{\boldsymbol{d}}^{(k+1)})_i - \left((\widehat{\boldsymbol{d}}^{(k)})_i - \beta^{(k-1)} \dfrac{(\widehat{\boldsymbol{d}}^{(k)})_i}{\left|(\widehat{\boldsymbol{d}}^{(k)})_i\right|}\right), & if \left|(\widehat{\boldsymbol{d}}^{(k)})_i\right| > \beta^{(k-1)} \\ (\widehat{\boldsymbol{d}}^{(k+1)})_i, & if \left|(\widehat{\boldsymbol{d}}^{(k)})_i\right| \leq \beta^{(k-1)} \end{cases} \tag{20}$$

and

$$\left(\epsilon_n^{(k)}\right)_i = \begin{cases} \beta^{(k-1)} \dfrac{(\widehat{\boldsymbol{d}}^{(k)})_i}{\left|(\widehat{\boldsymbol{d}}^{(k)})_i\right|}, & if \left|(\widehat{\boldsymbol{d}}^{(k)})_i\right| > \beta^{(k-1)} \\ (\widehat{\boldsymbol{d}}^{(k)})_i, & if \left|(\widehat{\boldsymbol{d}}^{(k)})_i\right| \leq \beta^{(k-1)}, \end{cases} \tag{21}$$

respectively. In similar lines to that of Lemma 3.8 in Daubechies *et al.* [34], enforcing sparsity by $l_1$-norm minimization leads to $\left|(\widehat{\boldsymbol{d}}^{(k+1)}) - (\widehat{\boldsymbol{d}}^{(k)})\right| \to 0$ as $k \to \infty$. The expectations of the two errors at convergence, can hence be approximated as

$$\mathbb{E}\left(\left|\epsilon_{\text{res}}^{(k+1)}\right|\right) \to \beta^{(k-1)} \mathbb{E}\left(\left|\widehat{\boldsymbol{d}}_p^{(k)}\right|\right) + \mathbb{E}(|\widehat{\boldsymbol{d}}^{(k+1)}|)_{<\beta^{(k-1)}}, \tag{22}$$

where $\widehat{\boldsymbol{d}}_p^{(k)} = \left[\dfrac{(\widehat{\boldsymbol{d}}^{(k)})_1}{|(\widehat{\boldsymbol{d}}^{(k)})_1|}, \cdots, \dfrac{(\widehat{\boldsymbol{d}}^{(k)})_{MN}}{|(\widehat{\boldsymbol{d}}^{(k)})_{MN}|}\right]$ and $\mathbb{E}(|\widehat{\boldsymbol{d}}^{(k+1)}|)_{<\beta^{(k-1)}}$ denotes the expectation of absolute gradients satisfying $|\widehat{\boldsymbol{d}}^{(k+1)}| < \beta^{(k-1)}$. In similar lines to (22),

$$\mathbb{E}\left(\left|\epsilon_n^{(k)}\right|\right) \to \beta^{(k-1)} \mathbb{E}\left(\left|\widehat{\boldsymbol{d}}_p^{(k)}\right|\right) + \mathbb{E}(|\widehat{\boldsymbol{d}}^{(k)}|)_{<\beta^{(k-1)}}, \tag{23}$$

where $\mathbb{E}(|\widehat{d}^{(k)}|)_{<\beta^{(k-1)}}$ denotes the expectation of absolute gradients satisfying $|\widehat{d}^{(k)}| < \beta^{(k-1)}$. Thus at convergence,

$$\mathbb{E}\left(\left|\epsilon_{\text{res}}^{(k+1)}\right|\right) - \mathbb{E}\left(\left|\epsilon_n^{(k)}\right|\right) = \mathbb{E}(|\widehat{d}^{(k+1)}|)_{<\beta^{(k-1)}} - \mathbb{E}(|\widehat{d}^{(k)}|)_{<\beta^{(k-1)}}. \tag{24}$$

Because the number of absolute gradients less than the desired threshold should decrease with iterations at convergence, it is expected that $\mathbb{E}(|\widehat{d}^{(k+1)}|)_{<\beta^{(k-1)}} > \mathbb{E}(|\widehat{d}^{(k)}|)_{<\beta^{(k-1)}}$. This confirms that $\mathbb{E}\left(\left|\epsilon_{\text{res}}^{(k+1)}\right|\right) > E\left(\left|\epsilon_n^{(k)}\right|\right)$. Also, since the threshold adaptation requires $\beta^{(k+1)} < \beta^{(k)}$, it is inferred that

$$\frac{\mathbb{E}\left(\left|\epsilon_{\text{res}}^{(k+1)}\right|\right)}{\beta^{(k+1)}} > \frac{\mathbb{E}\left(\left|\epsilon_n^{(k)}\right|\right)}{\beta^{(k)}}. \tag{25}$$

Taking the difference between the LHS and RHS as an increasing function $\Phi$ of $\mathfrak{d}_1^{(k)}$ satisfying the conditions specified in [31], the threshold update scheme takes the form

$$\beta^{(k+1)} = \frac{\mathbb{E}\left(\left|\epsilon_{\text{res}}^{(k+1)}\right|\right)}{\Phi\left(\mathfrak{d}_1^{(k)}\right) + \left(\mathbb{E}\left(\left|\epsilon_n^{(k)}\right|\right)/\beta^{(k)}\right)}. \tag{26}$$

The expected values can be computed from the histograms of $\left|\epsilon_{\text{res}}^{(k+1)}\right|$, $\left|\epsilon_n^{(k)}\right|$ and $\left|\epsilon_n^{(k-1)}\right|$. Thus if $\Phi(\cdot)$ is chosen as an increasing function of $\mathfrak{d}_1^{(k)}$ with $\Phi\left(\mathfrak{d}_1^{(k)}\right) \to 0$ as $\mathfrak{d}_1^{(k)} \to 0$, it is first required to numerically satisfy the limits for $\mathfrak{d}_1^{(k)}$ specified in [31]. We have used three sample functions $\mathfrak{d}_1^{(k)}$, $log\left(1 + \mathfrak{d}_1^{(k)}\right)$ or $1 - exp\left(-\mathfrak{d}_1^{(k)}\right)$ for $\Phi(\cdot)$ that satisfies the limits for all data sets used in our study. For other data sets, it may be useful to scale the functions as $\Phi\left(c\mathfrak{d}_1^{(k)}\right)$ and search for a suitable value of $c$ for which the limits are satisfied within the first few iterations.

*D. Application to Parallel MRI*

In each iteration, the derivative space errors, calculated from the sum-of-squares (SoS) image is utilized to obtain the threshold update as in the single channel case. The updated threshold is then applied to the individual channels for reconstruction. The steps for threshold adaptation are summarized below, with TVIS steps provided in (A1) for comparison. In (A2), the errors in derivative space, calculated from the SoS combined image, are utilized to obtain the update. The initialization steps are common to all algorithms, and include setting $d^{(0)} = \tilde{d}^{(0)} = 0$, $t^{(0)} = 1$, $tol = 1.0e - 4$, and $K_u$ as the zero-filled k-space with $n_C$ number of channels. The initial threshold $\beta^{(0)}$ is determined as outlined in section II.C.

**Algorithm-1 (A1) :** Standard TVIS with FISTA

*Iterate for $k = 1, \cdots, MaxIter$:*
  *1) Iterate for $c = 1, \cdots, n_C$:*
    *(i) Calculate $\epsilon_{res}^{(k)}$ as in (10)*
    *(ii) Find $\hat{d}^{(k)} = \tilde{d}^{(k-1)} + \epsilon_{res}^{(k)}$*
    *(iii) Soft Threshold the gradient domain Coefficients*
      $d^{(k)} = \mathcal{T}_\beta(\hat{d}^{(k)})$
    *(iv) Find $U_c^{(k)} = \mathcal{X}(d^{(k)})$*
    *(v) Find $t^{(k)} = 1 + \sqrt{1 + 4(t^{(k-1)})^2}$*
    *(vi) $\tilde{d}^{(k)} = d^{(k)} + \frac{t^{(k-1)} - 1}{t^{(k)}}(d^{(k)} - d^{(k-1)})$*
  *end*

  *2) Compute the SoS image $U^{(k)} = \sqrt{\sum_{c=1}^{n_C}(U_c^{(k)})^2}$*

  *3) Repeat steps (1) − (2) until $\frac{\|U^{(k)} - U^{(k-1)}\|_2}{\|U^{(k-1)}\|_2} \leq tol$.*
*end*

**Algorithm-2 (A2) :** ATVIS

*Iterate for $k = 1, \cdots, MaxIter$:*
 *1) Follow steps (1) − (2) as in ( **A1**)*
 *2) Calculate $\epsilon_n^{(k-1)}$ and $\epsilon_{res}^{(k)}$ from combined image*
 *3) Calculate the expectations of $|\epsilon_{res}^{(k)}|$ and $|\epsilon_n^{(k-1)}|$ from respective histograms*
 *4) Calculate $\eth_1^{(k)} = \|\epsilon_{res}^{(k)}\|_1 - \|\epsilon_n^{(k-1)}\|_1$*
 *5) Update $\beta^k$ using (26)*
 *6) Repeat steps (1) to (5) until $\frac{\|U^{(k)} - U^{(k-1)}\|_2}{\|U^{(k-1)}\|_2} \leq tol$.*
*end*

## III. RESULTS

*A. Application to image restoration*

In this section, we illustrate the application of the ATVIS algorithm to an image restoration problem, in which we attempt to estimate the unknown image from a noisy observed image, with a known system response function $H$. For an unknown image $U$, the noisy version $\breve{U}$ is modelled as

$$\breve{U} = HU + n, \qquad (27)$$

where $n$ represents the additive noise. The numerical simulations are performed using a Shepp Logan phantom ($256 \times 256$) at three different noise levels ($\sigma = 1e - 3$, $5e - 3$ and $1e - 2$) and two blurring kernels : Gaussian blur kernel ($G_k$) and Motion blur kernel ($M_k$). The degraded images are first obtained by convolving the phantom with the blurring kernel, followed by addition of ten different realizations of zero mean Gaussian white noise with standard deviation $\sigma$. In order to obtain statistically meaningful results for comparison between methods, the RLNE is averaged over all trials. The RLNE is computed using

$$RLNE = \|U - U_{\text{ref}}\|_2 / \|U_{\text{ref}}\|_2, \tag{28}$$

where $U_{\text{ref}}$ and $U$ denote the ground truth image and restored image, respectively. The RLNE values obtained using ATVIS and other restoration methods are summarized in Table I. For all methods, a common stopping criterion (as shown in the algorithm (A1) step-3) is employed. The algorithms are implemented using Matlab (The Mathworks, Nattick, MA) on a PC with Intel Xeon 2.4 GHz processor and 16 GB of RAM running Windows 7 OS. The adaptation function used is $\Phi\left(\mathfrak{d}_1^{(k)}\right) = \mathfrak{d}_1^{(k)}$. For purpose of reproducibility, all Matlab codes are made publicly available at http://www.iiitmk.ac.in/MedImagCompLab/codes/raji/atvis.rar. The restored images are shown in Fig. 1. The threshold for restoration using TVIS is computed from the threshold selection criterion defined in [32]. However, the initial threshold in the adaptive case is computed using the procedure defined in section II *C*. For faster implementation, over-relaxation is also included in both variants of TVIS algorithm as in FISTA [40, 41]. The left-most panel of Fig. 1 shows the input degraded image. Columns from left to right in the top row indicate images recovered using TVIS, derivative (D)-ADMM [42], adaptive parameter estimation (APE)-ADMM [25], and ATVIS, respectively. The difference images with RLNE values in the insets clearly indicate the superiority of ATVIS.

**Table I Here**

**Fig. 1 Here**

*B. Application to CS reconstruction*

*1) Numerical experiment*

The improved reconstruction performance of ATVIS in comparison to constant threshold TVIS is illustrated using a noise-free and noisy versions of two phantom images (I and II), undersampled with the sampling mask shown in Fig. 2. Each phantom image is reconstructed with TVIS and ATVIS. Plots of the threshold and RLNE versus iteration number are shown in Fig. 3. For all noise conditions, the TVIS reconstruction using a constant threshold yields slower and suboptimal reconstruction. The difference images of phantom I and II, with their RLNE values (indicated in insets) are shown in Figs. 4 and 5, respectively.

**Fig. 2 Here**

**Fig. 3 Here**

**Fig. 4 Here**

**Fig. 5 Here**

*2) Application to single channel data*

A fully sampled k-space of an image (obtained from http://www.quxiaobo.org) is retrospectively undersampled using a sampling mask shown in Fig. 2. The sampling mask is generated using 30% acquired samples, with 1.55% of samples fully acquired from the central k-space. The ATVIS reconstruction is compared with constant threshold TVIS. Top and bottom panels in Fig. 6 show the reconstructed and difference images. Columns from left to right show reconstructions using constant threshold TVIS and ATVIS, respectively, with their corresponding RLNE values provided in the insets.

**Fig. 6 Here**

*3) Application to multi-channel data*

The application of the proposed adaptation strategy to parallel MRI is illustrated using multi-channel in-vivo data. Raw k-space data were acquired from a GE Discovery 3T 750W clinical MR scanner at Sree Chitra Tirunal Institute of Medical Sciences and Technology, Trivandrum, India. All subjects were scanned with prior written informed consent as recommended by the institutional ethics committee. *Datasets*-I and II are T1-weighted structural images obtained using GE's BRAVO sequence. Fig. 7 illustrates reconstructions for *dataset*-I using TVIS, and ATVIS methods. The top two panels of Fig. 7 show the reconstructed and difference images obtained with a random sampling mask (MASK-1). The lower two panels correspond to a radial sampling scheme (MASK-2) with angular separation based on the golden ratio [43]. For the radial sampling scheme, we have used 80 spokes with 256 samples along each spoke. RLNE values shown in the insets indicate the improved quality of reconstruction obtained with the ATVIS method, as confirmed by visual inspection of the difference images.

**Fig.7 Here**

Fig. 8 illustrates reconstructions for *dataset*-II using the TVIS and ATVIS methods. The top two panels show the reconstructed and difference images obtained with a radial sampling scheme with uniform angular separation (MASK-3). Again, we have used 80 spokes with 256 samples along each spoke. The lower two panels correspond to a Cartesian sampled k-space, retrospectively undersampled along phase-encode direction based on the golden ratio (MASK-4) [44]. In the random phase-encode undersampling scheme, 120 phase encode lines are acquired. Of these, the 32 lines from the central region of k-space are equally spaced, and the rest are randomly spaced. RLNE values, as well as the initial and final thresholds, are shown in the insets. Visual inspection and RLNE values attest to the improved quality of ATVIS reconstruction.

**Fig. 8 Here**

*4) Application to dynamic MRI*

The adaptive scheme is applied to dynamic MRI in which several k-spaces are collected at multiple time points. As the accelerated dynamic MR images are more susceptible to artefacts, approaches that provide higher levels of sparsity, such as adaptive dictionary learning (DL), are preferred over wavelet or TV based schemes. In DL with temporal gradients (DLTG) [45], an additional sparsifying transform in the temporal dimension is used along with DL for improved dynamic MR image reconstruction of cardiac cine data. Here, we demonstrate the application of threshold adaptation to enforce sparsity of the temporal gradients. The implementation of DLTG is made up of three parts in an iterative framework, as outlined in Algorithm (A3) below. The input to the algorithm is a complex valued single-slice cardiac cine data denoted by $\mathbf{X}^{(3D)}$ and its concatenated column vector denoted by $\mathbf{x}$. The corresponding zero-filled sequence is denoted by $\mathbf{x}_z$. In DL, $\mathbf{x}_{T,i}$ denotes the training patches represented as column vectors of size $m$ and $\Gamma_T$ is the matrix formed with columns $\gamma_{T,i}$, each containing the sparse representation of $\mathbf{x}_{T,i}$. The dictionary is denoted as $\boldsymbol{D}$, with real and imaginary parts denoted as $\Re(\mathbf{x})$ and $\Im(\mathbf{x})$, and their corresponding sparse coding as $\Gamma_\Re$ and $\Gamma_\Im$, respectively. With $R_i$ denoting an operator that extracts a column vector corresponding to a 3D patch in the dataset starting from pixel location $i$, the columns of $\Gamma_\Re$ and $\Gamma_\Im$ (represented as $\gamma_{\Re,i}$ and $\gamma_{\Im,i}$) are obtained by the coding of patch $R_i\Re(\mathbf{x})$ and $R_i\Im(\mathbf{x})$, respectively.

---

**Algorithm-3 (A3) : DLTG**

***Initialization***
*Input the zero-filled sequence* $\mathbf{x}_z$.
*Initialize* $r_1 = 0, r_2 = 0, \mathbf{x}^{(0,0)} = \mathbf{x}_z, \Gamma_\Re = \Gamma_\Im = 0, tol = 1.0e-4$.
***Main Iteration***

1) *Iterate for* $r_1 = 1, \cdots, MaxIter1$:
   (i) *Dictionary Training*
      $\{\boldsymbol{D}, \Gamma_T\} \leftarrow \min_{\boldsymbol{D},\Gamma_T} \|\gamma_{T,i}\|_0 \ s.t. \ \|R_i \mathbf{x}_T^{(r_1,r_2)} - \boldsymbol{D}\gamma_{T,i}\|_2^2 < \epsilon$
   (ii) *Sparse Coding of Real and Imaginary parts separately*
      $\Gamma_\Re \leftarrow \min_{\Gamma_\Re} \|\gamma_{R,i}\|_0 \ s.t. \ \|R_i \Re(\mathbf{x}^{(r_1,r_2)}) - \boldsymbol{D}\gamma_{\Re,i}\|_2^2 < \epsilon$
      $\Gamma_\Im \leftarrow \min_{\Gamma_\Im} \|\gamma_{I,i}\|_0 \ s.t. \ \|R_i \Im(\mathbf{x}^{(r_1,r_2)}) - \boldsymbol{D}\gamma_{\Im,i}\|_2^2 < \epsilon$
   (iii) *DL Sparse Approximation*
      $U' \leftarrow \frac{1}{n} \sum_{i=1}^{P} R_i^T \boldsymbol{D}(\gamma_{\Re,i} + \gamma_{\Im,i})$
   (iv) *Iterate for* $r_2 = 1, \cdots, MaxIter1$:
      (a) *Enforce data consistency*
      (b) *Enforce TG Sparsity*
      $\mathbf{x}' = \min_{\mathbf{x}} \|\mathbf{x}^{(r_1,r_2)} - \mathbf{x}'\|_2^2 + \beta^{(r_2)} \|\nabla_t |\mathbf{x}'|\|_1$
      end
   (v) *Repeat steps* (i) − (iv) *until* $\frac{\|\mathbf{x}^{(r_1,r_2)} - \mathbf{x}^{(pr\,1,pr\,2)}\|_2}{\|\mathbf{x}^{(pr\,1,pr\,2)}\|_2} \leq tol$, *where* $\mathbf{x}^{(pr_1,pr_2)}$ *is the previous iterate of the solution.*
end

The adaptive implementation of DLTG follows the threshold $\beta$ update scheme in (26). The use of temporal gradient sparsity in DLTG affects the convergence rate of the algorithm, as shown in [45]. Setting the threshold parameter to a high initial value and reducing it with succeeding iterations results in faster convergence and lower steady-state errors as compared to DLTG with a constant parameter. Thus, the adaptation scheme applied to temporal gradient thresholding can accelerate the convergence, while yielding lower steady-state reconstruction errors. Fig. 9 depicts reconstructions of dynamic data (obtained from http://caballerojose.com/code.html.) containing 30 temporal frames of size $256 \times 256$, obtained using the constant parameter and the adaptive parameter for enforcing sparsity along the temporal dimension. The top two panels of Fig. 9 show the reconstructed and difference images obtained with the sampling scheme labeled MASK-5. The lower two panels include the temporal profile of row 128 in the original dataset and their respective difference images.

**Fig. 9 Here**

IV. DISCUSSION

In this paper, a threshold adaptation strategy for a TV iterative shrinkage scheme is proposed, in order to achieve a faster reduction of reconstruction error and convergence to the minimum possible steady-state error. Similar to GDP which dictates the non-negativity of difference between $l_2$-norms of consistency and perturbation errors, the proposed adaptation scheme entails reduction of the discrepancy level in TV norms of the consistency and sparse approximation errors estimated in the derivative space. The absolute difference between the TV norms of the consistency and sparse approximation errors, together with the expectations of their respective spatial distributions, is required for determination of the threshold.

In estimating the missing samples in compressed sensing MRI, regularization provides a mechanism for balancing the two errors. The presence of noise or artefacts in the resultant image is attributed to the increase in either of these two errors. Thus, low consistency errors obtained by relaxing the difference domain sparsity will result in a noisy reconstruction. Conversely, relaxing the model, along with excess truncation in the difference domain, can result in artefact-prone reconstruction. In summary, the method works iteratively to improve image quality by continuously reducing the consistency and sparse approximation errors. The speed of reconstruction depends on how fast the errors decrease in relation to each other. Further ramifications of key observations of the interplay between the two types of errors could be applied to other inverse imaging problems.

As the TVIS algorithm is a single-step shrinkage scheme, its rate of convergence can be improved by the incorporation of two-step IS strategies, such as TwIST [46] and FISTA [40, 41] algorithms. The time advantage of the adaptation schemes becomes more relevant when used for multi-channel reconstructions, since they involve multiple coil reconstructions in a single iteration. Compute times observed in the multi-channel reconstructions, demonstrate that the computational cost of the ATVIS algorithm is 30% lower than that of FISTA.

Furthermore, the proposed scheme is more suited to practical implementation in pMRI and dynamic MRI, since the adaptation does not require computation of the exact upper bound on the perturbation error, which requires accurate knowledge of the measurement or input noise as in the discrepancy principle. Other parameter adaptation schemes in TV based image restoration, such as APE-ADMM, require prior knowledge of noise, thus limiting the practical use of such discrepancy-based algorithms. The proposed adaptation scheme is shown to result in reduction of reconstruction error, along with acceleration of the compute time. Further improvements in image quality can be achieved using higher-order TV based implementations, referred to as the total generalized variation (TGV) [47, 48]. TGV was introduced to control or eliminate the staircasing effect that often manifests in reconstructions using first-order TV implementations. The staircasing artefacts occur because the piece-wise constant assumption required for efficient TV recovery in MR image reconstruction, is not satisfied due to the inhomogeneities of the exciting **B1** field of high field systems (3.0 T and above), and of the receive coils. Although the present work is restricted to adaptation of the first-order TV regularization parameter, it can be easily extended to higher-order TV implementation as a future extension of this work.

## V. CONCLUSION

This paper establishes an adaptive thresholding scheme for the TVIS algorithm to accelerate sparse image reconstruction for compressed sensing MRI. The acceleration is attained by a fast minimization of the absolute difference between TV norms of the consistency and sparse approximation errors in derivative space. The rapid norm minimization is achieved by introducing a varying threshold in each iteration, updated adaptively using statistical measures derived from criteria similar to GDP.

## ACKNOWLEDGEMENT

The authors wish to acknowledge Prof. Frithjof Kruggel, Department of Biomedical Engineering, University of California, Irvine, Prof. Michael Braun, University of Technology, Sydney and Prof. Socrates Dokos, Graduate school of Biomedical Engineering, The University of New South Wales, Sydney, for their patience to proofread this manuscript. The authors are also grateful to Chuan He, High-Tech Institute of Xi'an, Xi'an, China and Dongwei Ren, School of Computer Science and Technology, Harbin Institute of Technology, Harbin, China for making their APE-ADMM and D-ADMM codes accessible.

## REFERENCES

[1] M. Lustig, D. Donoho, and J. M. Pauly, "Sparse MRI: The application of compressed sensing for rapid MR imaging," *Magnetic resonance in medicine,* vol. 58, no. 6, pp. 1182-1195, 2007.
[2] A. Chambolle, R. A. De Vore, N.-Y. Lee, and B. J. Lucier, "Nonlinear wavelet image processing: variational problems, compression, and noise removal through wavelet shrinkage," *IEEE Transactions on Image Processing,* vol. 7, no. 3, pp. 319-335, 1998.
[3] M. Fornasier and H. Rauhut, "Iterative thresholding algorithms," *Applied and Computational Harmonic Analysis,* vol. 25, no. 2, pp. 187-208, 2008.
[4] M. A. Figueiredo and R. D. Nowak, "An EM algorithm for wavelet-based image restoration," *IEEE Transactions on Image Processing,* vol. 12, no. 8, pp. 906-916, 2003.
[5] X. Qu, W. Zhang, D. Guo, C. Cai, S. Cai, and Z. Chen, "Iterative thresholding compressed sensing MRI based on contourlet transform," *Inverse Problems in Science and Engineering,* vol. 18, no. 6, pp. 737-758, 2010.
[6] E. Candes, L. Demanet, D. Donoho, and L. Ying, "Fast discrete curvelet transforms," *Multiscale Modeling & Simulation,* vol. 5, no. 3, pp. 861-899, 2006.
[7] L. I. Rudin, S. Osher, and E. Fatemi, "Nonlinear total variation based noise removal algorithms," *Physica D: nonlinear phenomena,* vol. 60, no. 1-4, pp. 259-268, 1992.
[8] A. Chambolle and P.-L. Lions, "Image recovery via total variation minimization and related problems," *Numerische Mathematik,* vol. 76, no. 2, pp. 167-188, 1997.
[9] I. Daubechies, G. Teschke, and L. Vese, "Iteratively solving linear inverse problems under general convex constraints," *Inverse Problems and Imaging,* vol. 1, no. 1, p. 29, 2007.
[10] D. Strong and T. Chan, "Edge-preserving and scale-dependent properties of total variation regularization," *Inverse problems,* vol. 19, no. 6, p. S165, 2003.

[11] C. R. Vogel and M. E. Oman, "Iterative methods for total variation denoising," *SIAM Journal on Scientific Computing,* vol. 17, no. 1, pp. 227-238, 1996.
[12] T. F. Chan, G. H. Golub, and P. Mulet, "A nonlinear primal-dual method for total variation-based image restoration," *SIAM journal on scientific computing,* vol. 20, no. 6, pp. 1964-1977, 1999.
[13] A. Chambolle, "An algorithm for total variation minimization and applications," *Journal of Mathematical imaging and vision,* vol. 20, no. 1-2, pp. 89-97, 2004.
[14] A. Chambolle and T. Pock, "A first-order primal-dual algorithm for convex problems with applications to imaging," *Journal of mathematical imaging and vision,* vol. 40, no. 1, pp. 120-145, 2011.
[15] Y. Wang, J. Yang, W. Yin, and Y. Zhang, "A new alternating minimization algorithm for total variation image reconstruction," *SIAM Journal on Imaging Sciences,* vol. 1, no. 3, pp. 248-272, 2008.
[16] T. Goldstein and S. Osher, "The split Bregman method for L1-regularized problems," *SIAM journal on imaging sciences,* vol. 2, no. 2, pp. 323-343, 2009.
[17] M. K. Ng, P. Weiss, and X. Yuan, "Solving constrained total-variation image restoration and reconstruction problems via alternating direction methods," *SIAM journal on Scientific Computing,* vol. 32, no. 5, pp. 2710-2736, 2010.
[18] M. Carlavan and L. Blanc-Féraud, "Sparse Poisson noisy image deblurring," *IEEE Transactions on Image Processing,* vol. 21, no. 4, pp. 1834-1846, 2012.
[19] T. Teuber, G. Steidl, and R. H. Chan, "Minimization and parameter estimation for seminorm regularization models with I-divergence constraints," *Inverse Problems,* vol. 29, no. 3, p. 035007, 2013.
[20] M. Burger, A. Sawatzky, and G. Steidl, "First order algorithms in variational image processing," in *Splitting Methods in Communication, Imaging, Science, and Engineering*: Springer, 2016, pp. 345-407.
[21] R. Glowinski, S. J. Osher, and W. Yin, *Splitting Methods in Communication, Imaging, Science, and Engineering*. Springer, 2017.
[22] V. A. Morozov, *Methods for solving incorrectly posed problems*. Springer Science & Business Media, 2012.
[23] Y.-W. Wen and R. H. Chan, "Parameter selection for total-variation-based image restoration using discrepancy principle," *IEEE Transactions on Image Processing,* vol. 21, no. 4, pp. 1770-1781, 2012.
[24] Y.-W. Wen and A. M. Yip, "Adaptive parameter selection for total variation image deconvolution," *Numer. Math. Theor. Meth. Appl,* vol. 2, no. 4, pp. 427-438, 2009.
[25] C. He, C. Hu, W. Zhang, and B. Shi, "A fast adaptive parameter estimation for total variation image restoration," *IEEE Transactions on Image Processing,* vol. 23, no. 12, pp. 4954-4967, 2014.
[26] H. Liao, F. Li, and M. K. Ng, "Selection of regularization parameter in total variation image restoration," *JOSA A,* vol. 26, no. 11, pp. 2311-2320, 2009.
[27] Y. Lin, B. Wohlberg, and H. Guo, "UPRE method for total variation parameter selection," *Signal Processing,* vol. 90, no. 8, pp. 2546-2551, 2010.
[28] S. D. Babacan, R. Molina, and A. K. Katsaggelos, "Parameter estimation in TV image restoration using variational distribution approximation," *IEEE transactions on image processing,* vol. 17, no. 3, pp. 326-339, 2008.
[29] S. D. Babacan, R. Molina, and A. K. Katsaggelos, "Variational Bayesian blind deconvolution using a total variation prior," *IEEE Transactions on Image Processing,* vol. 18, no. 1, pp. 12-26, 2009.
[30] J. P. Oliveira, J. M. Bioucas-Dias, and M. A. Figueiredo, "Adaptive total variation image deblurring: a majorization–minimization approach," *Signal Processing,* vol. 89, no. 9, pp. 1683-1693, 2009.
[31] R. S. Mathew and J. S. Paul, "Sparsity Promoting Adaptive Regularization for Compressed Sensing Parallel MRI," *IEEE Transactions on Computational Imaging,* vol. 4, no. 1, pp. 147-159, 2018.
[32] O. V. Michailovich, "An iterative shrinkage approach to total-variation image restoration," *IEEE Transactions on Image Processing,* vol. 20, no. 5, pp. 1281-1299, 2011.
[33] B. K. Natarajan, "Sparse approximate solutions to linear systems," *SIAM journal on computing,* vol. 24, no. 2, pp. 227-234, 1995.
[34] I. Daubechies, M. Defrise, and C. De Mol, "An iterative thresholding algorithm for linear inverse problems with a sparsity constraint," *Communications on pure and applied mathematics,* vol. 57, no. 11, pp. 1413-1457, 2004.
[35] Y. Liu, Z. Zhan, J.-F. Cai, D. Guo, Z. Chen, and X. Qu, "Projected iterative soft-thresholding algorithm for tight frames in compressed sensing magnetic resonance imaging," *IEEE transactions on medical imaging,* vol. 35, no. 9, pp. 2130-2140, 2016.
[36] X. Qu *et al.*, "Undersampled MRI reconstruction with patch-based directional wavelets," *Magnetic resonance imaging,* vol. 30, no. 7, pp. 964-977, 2012.
[37] X. Qu, Y. Hou, F. Lam, D. Guo, J. Zhong, and Z. Chen, "Magnetic resonance image reconstruction from undersampled measurements using a patch-based nonlocal operator," *Medical image analysis,* vol. 18, no. 6, pp. 843-856, 2014.
[38] S. Sardy and H. Monajemi, "Threshold Selection for Total Variation Denoising," *arXiv preprint arXiv:1605.01438,* 2016.
[39] D. L. Donoho and J. M. Johnstone, "Ideal spatial adaptation by wavelet shrinkage," *biometrika,* vol. 81, no. 3, pp. 425-455, 1994.
[40] A. Beck and M. Teboulle, "A fast iterative shrinkage-thresholding algorithm for linear inverse problems," *SIAM journal on imaging sciences,* vol. 2, no. 1, pp. 183-202, 2009.

[41]   A. Beck and M. Teboulle, "Fast gradient-based algorithms for constrained total variation image denoising and deblurring problems," *IEEE Transactions on Image Processing,* vol. 18, no. 11, pp. 2419-2434, 2009.
[42]   D. Ren, H. Zhang, D. Zhang, and W. Zuo, "Fast total-variation based image restoration based on derivative alternated direction optimization methods," *Neurocomputing,* vol. 170, pp. 201-212, 2015.
[43]   L. Feng *et al.*, "Golden-angle radial sparse parallel MRI: Combination of compressed sensing, parallel imaging, and golden-angle radial sampling for fast and flexible dynamic volumetric MRI," *Magnetic resonance in medicine,* vol. 72, no. 3, pp. 707-717, 2014.
[44]   Y. Yang, F. Liu, Z. Jin, and S. Crozier, "Aliasing artefact suppression in compressed sensing MRI for random phase-encode undersampling," *IEEE Transactions on Biomedical Engineering,* vol. 62, no. 9, pp. 2215-2223, 2015.
[45]   J. Caballero, A. N. Price, D. Rueckert, and J. V. Hajnal, "Dictionary learning and time sparsity for dynamic MR data reconstruction," *IEEE transactions on medical imaging,* vol. 33, no. 4, pp. 979-994, 2014.
[46]   J. M. Bioucas-Dias and M. A. Figueiredo, "A new TwIST: Two-step iterative shrinkage/thresholding algorithms for image restoration," *IEEE Transactions on Image processing,* vol. 16, no. 12, pp. 2992-3004, 2007.
[47]   K. Bredies, K. Kunisch, and T. Pock, "Total generalized variation," *SIAM Journal on Imaging Sciences,* vol. 3, no. 3, pp. 492-526, 2010.
[48]   F. Knoll, K. Bredies, T. Pock, and R. Stollberger, "Second order total generalized variation (TGV) for MRI," *Magnetic resonance in medicine,* vol. 65, no. 2, pp. 480-491, 2011.

## Figures

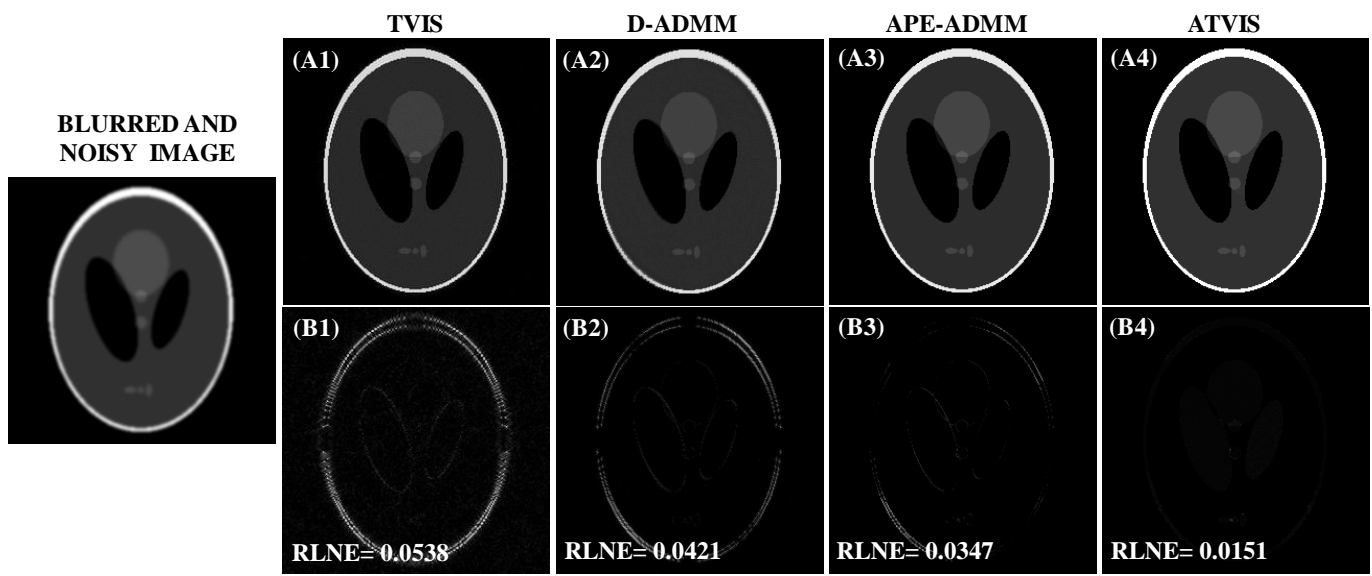

**Fig. 1: Image restoration using the ATVIS algorithm. Left-most panel shows the input degraded image. Columns from left to right in the top row show the images recovered using TVIS, D-ADMM, APE-ADMM and ATVIS algorithms, respectively. The lower panels show difference images, with the RLNE values indicated in the insets.**

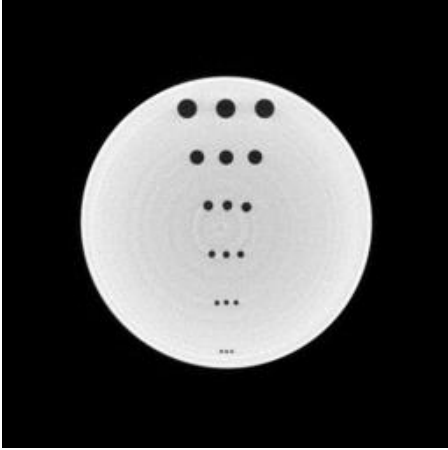 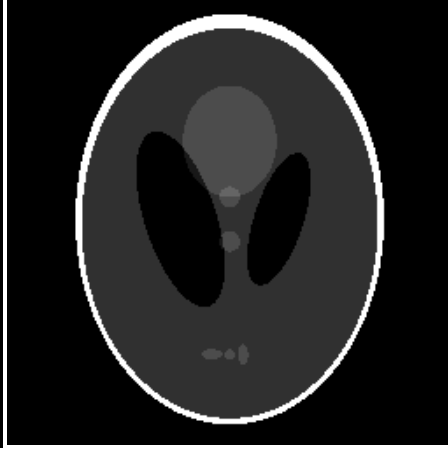 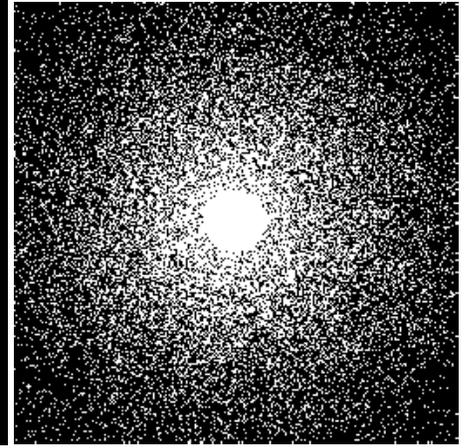

**Fig. 2: (A)-(B) Phantom images, (C) The sampling mask generated using 30% acquired samples, with 1.55% of samples fully acquired from the central k-space.**

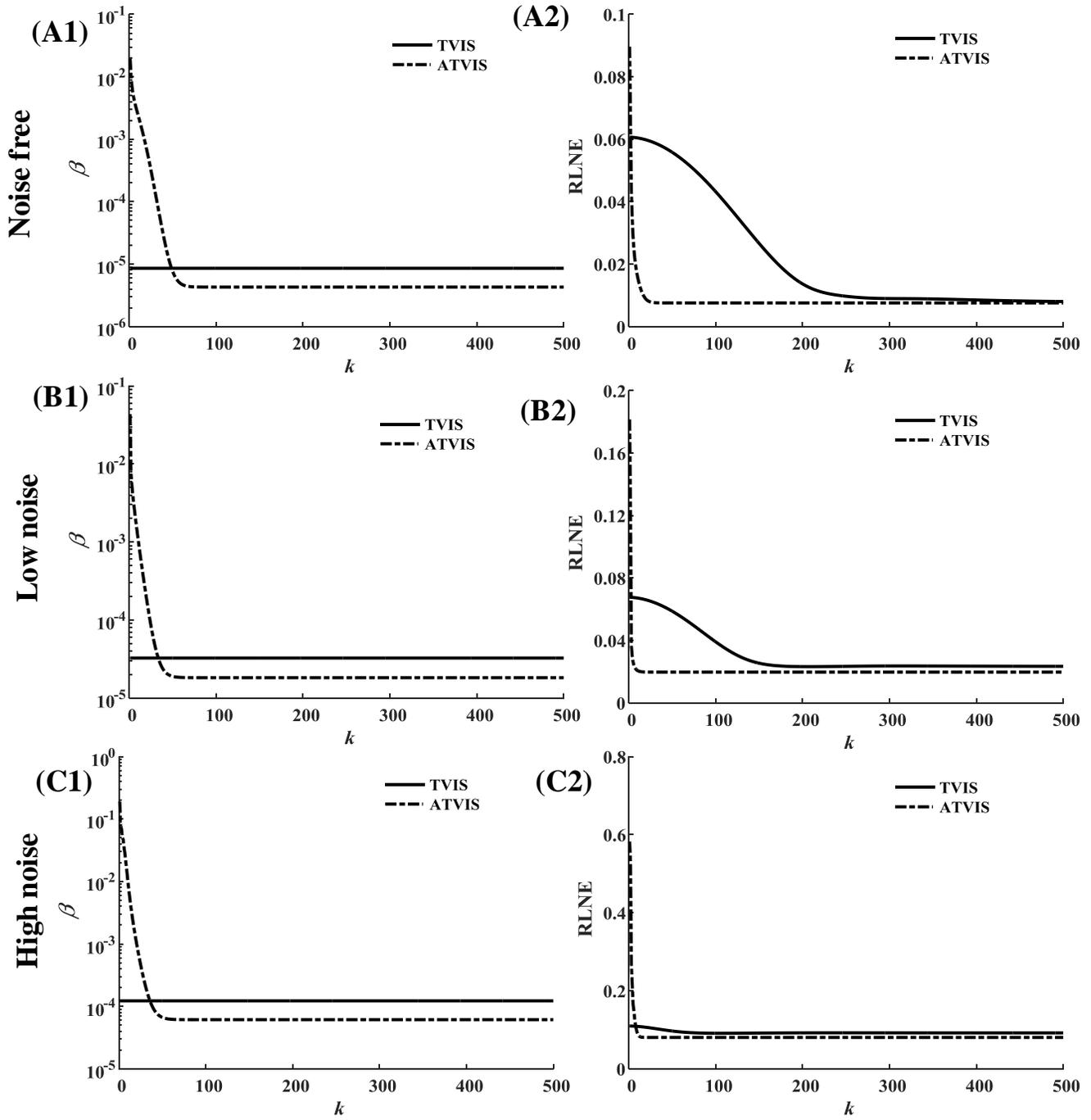

**Fig. 3:** Plots of threshold (left) and RLNE (right) as functions of iteration number at three different noise levels.

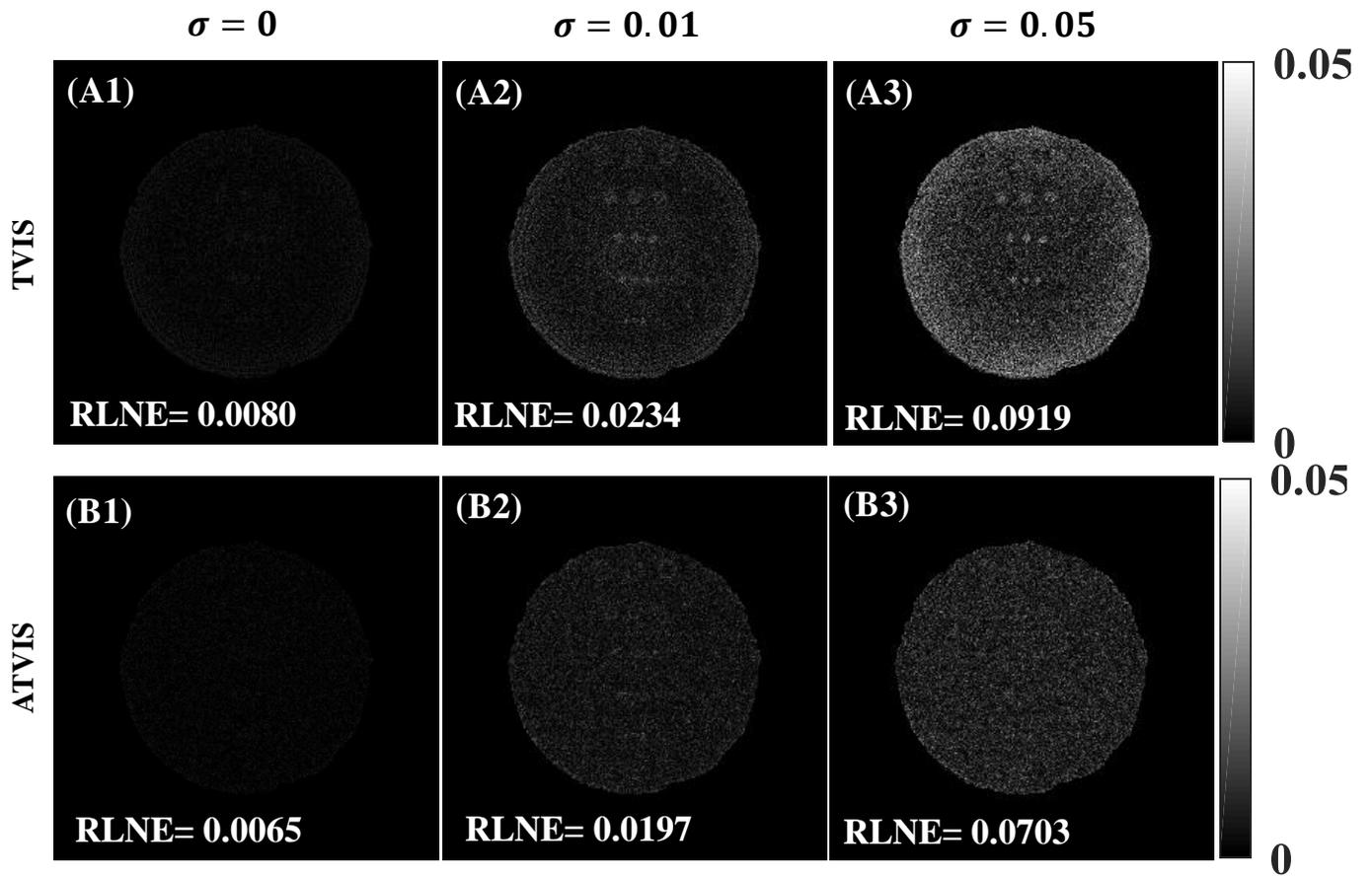

Fig. 4: Difference images of TVIS and ATVIS reconstructions using phantom-I for three different input noise levels.

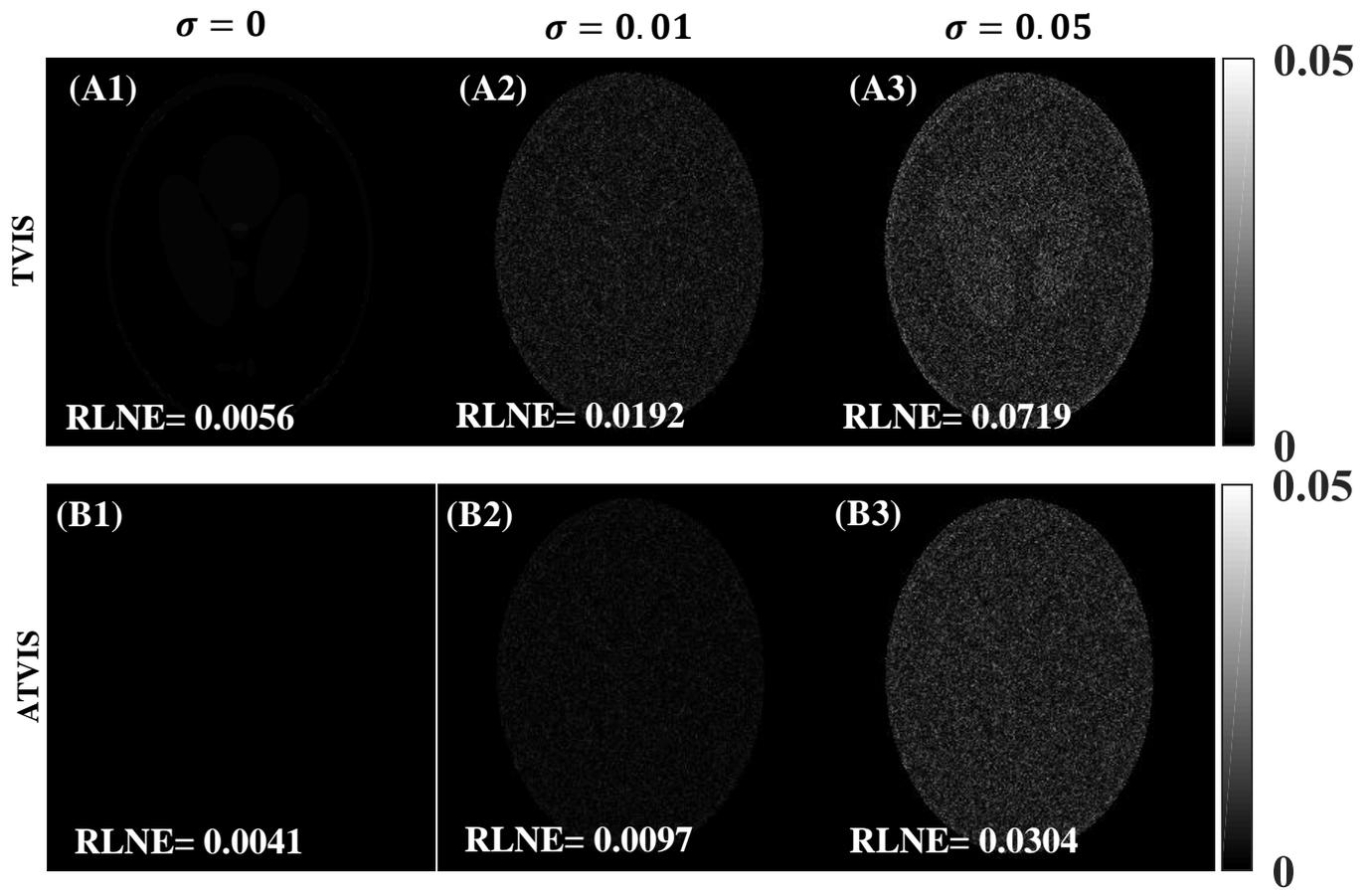

Fig. 5: Difference images of TVIS and ATVIS reconstructions using phantom-II for three different input noise levels.

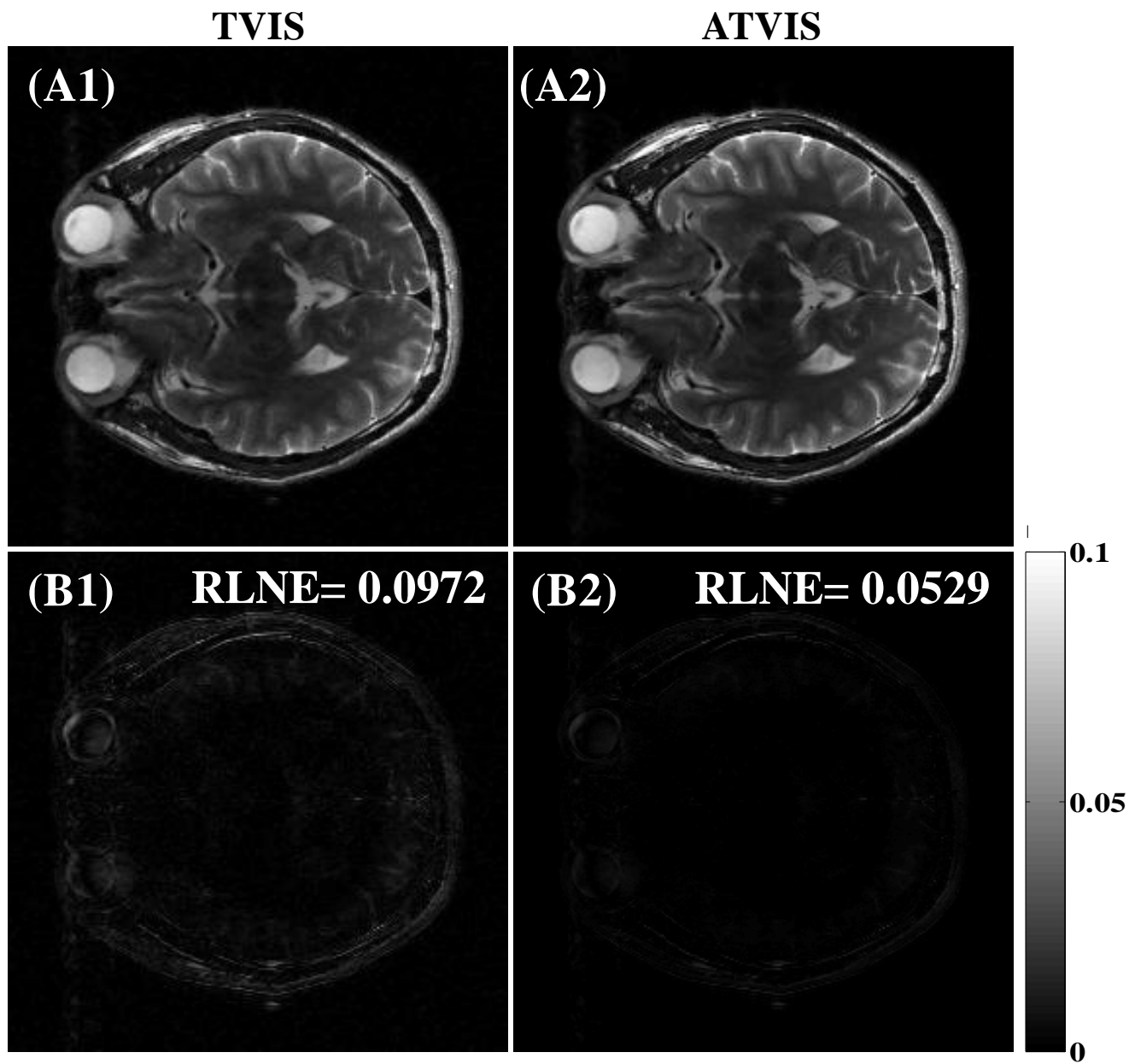

Fig. 6: Single channel reconstructions using (A1) TVIS, and (A2) ATVIS. (B1-B2) Corresponding difference images.

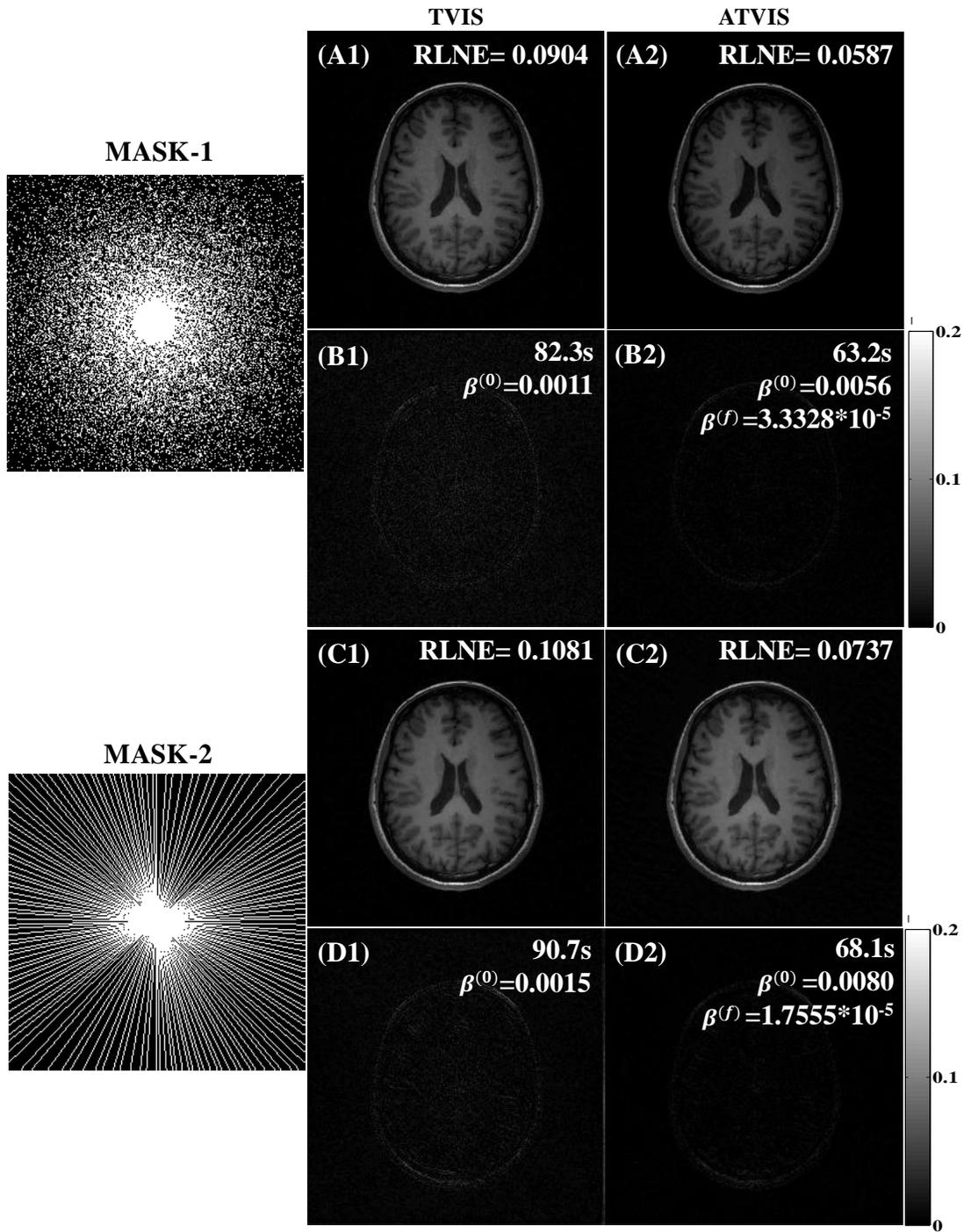

Fig. 7: Multi-channel reconstructions for *dataset*-I using (A1 and C1) TVIS, and (A2 and C2) ATVIS. Top and bottom panels show reconstructed and difference images for retrospective undersampling using MASKs-1 and 2, respectively. The insets in the difference image depict the reconstruction time, initial threshold ($\beta^{(0)}$) and final threshold ($\beta^{(f)}$) values. For TVIS, the initial and final thresholds are the same.

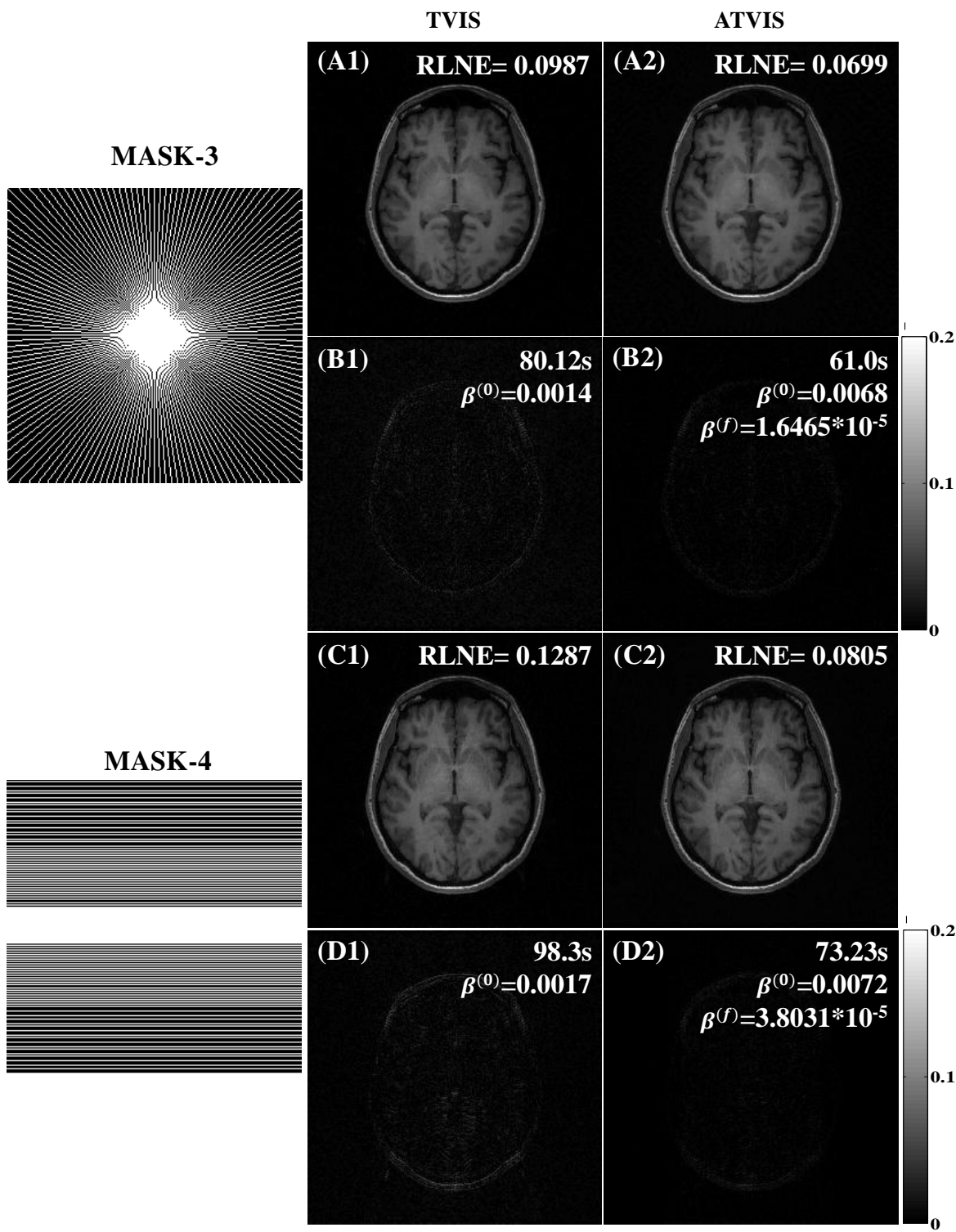

**Fig. 8:** Multi-channel reconstructions for *dataset*-II using (A1 and C1) TVIS, and (A2 and C2) ATVIS. Top and bottom panels show reconstructed and difference images for retrospective undersampling using MASKs-3, and 4, respectively. The insets in the difference image depict the reconstruction time, initial threshold ($\beta^{(0)}$) and final threshold ($\beta^{(f)}$) values. For TVIS, the initial and final thresholds are the same.

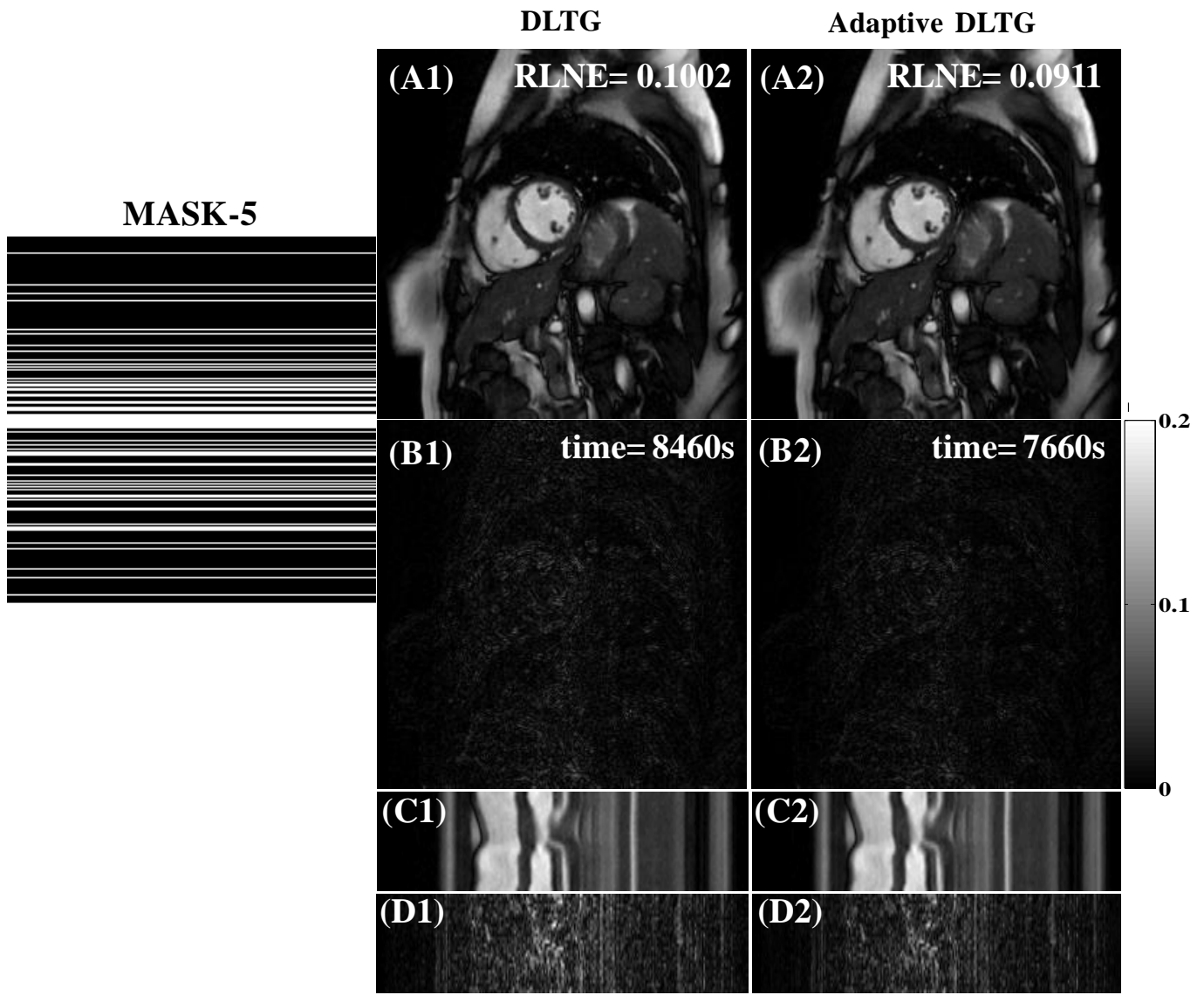

Fig. 9: Dynamic MRI reconstructions for cardiac cine data using (A1) DLTG, and (A2) adaptive DLTG. The top and bottom panels show reconstructed and difference images for retrospective undersampling using MASK-5. The lower two panels show the temporal profile of row 128 in the original dataset and their respective difference images amplified by 6.

TABLE I

COMPARISON IN RLNE OF THE IMAGE DEBLURRING EXPERIMENT

| $\sigma$ | kernel | TVIS | D-ADMM | APE-ADMM | ATVIS |
|---|---|---|---|---|---|
| $1e-3$ | $G_k$ | 0.0538 | 0.0421 | 0.0347 | 0.0151 |
| | $M_k$ | 0.0380 | 0.0511 | 0.0225 | 0.0199 |
| $5e-3$ | $G_k$ | 0.0745 | 0.0564 | 0.0481 | 0.0194 |
| | $M_k$ | 0.0825 | 0.0805 | 0.0430 | 0.0299 |
| $1e-2$ | $G_k$ | 0.0906 | 0.1021 | 0.0882 | 0.0429 |
| | $M_k$ | 0.0912 | 0.1001 | 0.0792 | 0.0690 |